\documentclass[10pt,twocolumn,letterpaper]{article}

\usepackage{iccv}
\usepackage{times}
\usepackage{epsfig}
\usepackage{graphicx}
\usepackage{amsmath}
\usepackage{amssymb}
\usepackage{subfig}
\usepackage{authblk}

\usepackage[pagebackref=true,breaklinks=true,letterpaper=true,colorlinks,bookmarks=false]{hyperref}

\iccvfinalcopy 

\begin{document}

\title{VITAL: A Visual Interpretation on Text with Adversarial Learning for Image Labeling}

\author[1,3]{Tao Hu}
\author[2]{Chengjiang Long\thanks{This work was co-supervised by Chengjiang Long and Chunxia Xiao.}}
\author[1]{Leheng Zhang}
\author[1]{Chunxia Xiao}
\affil[1]{School of Computer Science, Wuhan University, Wuhan, Hubei, China}
\affil[2]{Kitware Inc. Clifton Park, NY, USA}
\affil[3]{School of Information Engineering, Hubei University for Nationalities, Enshi, Hubei, China}
\affil[ ]{\tt\small {hutao\_es@foxmail.com, chengjiang.long@kitware.com, 375602133@qq.com, cxxiao@whu.edu.cn}}
\maketitle

\begin{abstract}
   In this paper, we propose a novel way to interpret text information by extracting visual feature presentation from multiple high-resolution and photo-realistic synthetic images generated by Text-to-image Generative Adversarial Network (GAN) to improve the performance of image labeling. Firstly, we design a stacked Generative Multi-Adversarial Network (GMAN), StackGMAN++, a modified version of the current state-of-the-art Text-to-image GAN, StackGAN++,  
to generate multiple synthetic images with various prior noises conditioned on a text. And then we extract deep visual features from the generated synthetic images to explore the underlying visual concepts for text. Finally, we combine image-level visual feature, text-level feature and visual features based on synthetic images together to predict labels for images. We conduct experiments on two benchmark datasets, i.e., the Oxford 102 Category Flower Dataset  
and the Caltech-UCSD Birds-200-2011 Dataset，
and the experimental results clearly demonstrate the efficacy of our proposed approach.
\end{abstract}

\section{Introduction}

Nowadays, images are being taken and shared to be commented at an unprecedented rate among social networks like Facebook, Twitter, and Flickr. To help users efficiently organize and manage such media data from a very huge collection, it is necessary and practical to collect labeled visual datasets at large scale to develop automatic tools with robust machine learning approaches~\cite{long2013active, hua2013collaborative, long2015multi, long2016joint, Long_2017_CVPR, hua2018collaborative}. However, most of the current annotation platforms like Amazon Mechanical Turk~\cite{Crowston:2012amazon} and LabelMe~\cite{Russell:2008labelme} do not make full use of the text information to aid the labeling problems, although most images on the web carry rich text information which includes informative and semantic signals like who took the photo and, where and with whom. 

\begin{figure}
\begin{center}
\includegraphics[height=0.70\linewidth, width=0.99\linewidth, angle=0]{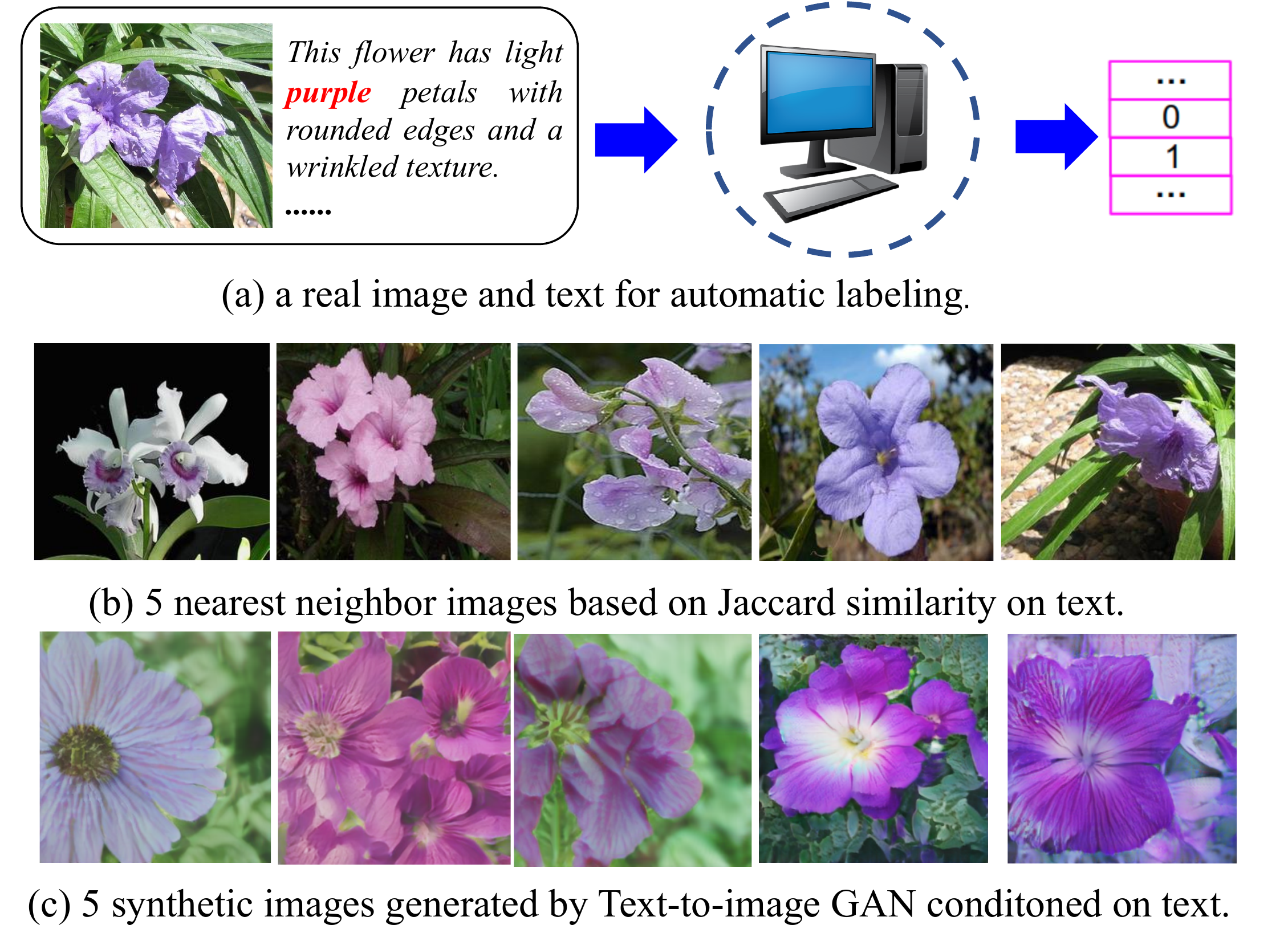}
\end{center}
\vspace{-0.40cm}
\caption{Illustration of two ways to visually interpret text for automatic image labeling (a). One is Johnson {\em et al.}'s work~\cite{Johnson:ICCV2015} in which a group of nearest neigbor images (b) defined based on Jaccard similiarity between image metadata especially tags extracted from a text. And the other one is our proposed VITAL method, which makes full use of $K$ high-resolution and photo-realist synthetic images (c) generated from StackGMAN++ (our modified version on StackGAN++~\cite{zhang:Arxiv2017}) conditioned on text. The goal of this paper is to visually interprete text information and extract visual feature representations from synthetic images to boost the accuracy of image labeling on real images.}
\label{fig:introduction}
\end{figure}

Prior work takes advantage of text context information to improve image classification by various treatments like selecting top frequently used words and user-generated tags~\cite{McAuleyECCV12, Hu:CVPR2016learning, Liu:CVPR2017semantic}, extracting text-level feature representation with Text Convolutional Neural Networks (CNNs)~\cite{Long:Arxiv2018}, and exploring visual feature representation from a set of neighbor images~\cite{Johnson:ICCV2015} defined based on Jaccard similarity between image metadata especially tags extracted from a text. The intuition behind is that images with similar text context information tend to depict similar scenes.

Inspired by the development of Text-to-image Generative Adversarial Networks (GANs)~\cite{Nguyen:CVPR2017, zhang:Arxiv2017stackgan, zhang:Arxiv2017, Xu:CVPR2018}, which is able to generate high-resolution and photo-realistic synthetic images conditioned on a text, we propose a novel visual interpretation on text with adversarial learning, named as ``VITAL", by interpreting text information with visual concepts extracted from a series of visually plausible synthetic images generated by Text-to-image GANs. 

Unlike Johnson {\em et al.}'s work~\cite{Johnson:ICCV2015} which uses a set of nearest neighbor images based on Jaccard similarity bewteen the text of query image and texts of training images, our proposed VITAL approach generates synthetic images conditioned on only the text of query image without using text from any other images.
As illustrated in Figure~\ref{fig:introduction}, the color information of nearest neighbor images to represent text are not well consistent with the real image, while synthetic images generated in our VITAL are able to not only cover most of the content in text information, but also provide much information about the background which is underlying in text information. This is also consistent with our human understanding to text information. 

``As there are 1000 Hamlets, there are 1000 readers." Usually, given a text that describes a specific scene, different readers can image different relevant visual scenes in their brains. Obviously, one synthetic image is not sufficient to simulate what multiple readers can visually interpret from a single text information. In principle, any text-to-
image GANs can be extended and incorporated into our VITAL framework and in this paper we just start to extend one current state-of-the-art Text-to-image GAN, StackGAN++~\cite{zhang:Arxiv2017}, dubbed StackGMAN++ in brief with ``M" indicating multiple generators, to extend the number of generators from the original $1$ to $K$ with different noise priors at each stage. The intuition behind is that each reader interprets Hamlets based on his/her own prior knowledge, and different prior knowledge leads to a different image of Hamlet in his/her mind. 

With $K$ generators, we are able to generate $K$ synthetic images. Since ResNet~\cite{HeResnet:CVPR2016} has demonstrated successes in a vast of vision applications, we apply it to extract visual feature for each synthetic image. It worth emphasizing that we care much more about the common visual representation among these $K$ synthetic images rather than each individual. Therefore, to obtain a compact visual feature representation, we first apply an affine transformation with a ReLU layer to adjust feature maps and reduce the channels before we apply an element-wise pooling to extract the common feature representation. 

Considering that feature fusion has been proved to be able to effectively improve the performance of image labeling, we combine the real image feature extracted from an image-level CNN, text feature extracted from a text-level CNN, and the common feature representation from $K$ synthetic images by concatenation and feed them into a fully connected layer as a classification for image labeling~\cite{Niu:TIP2019multi}.

\begin{figure*}
\begin{center}
\includegraphics[width=0.90\linewidth, angle=0]{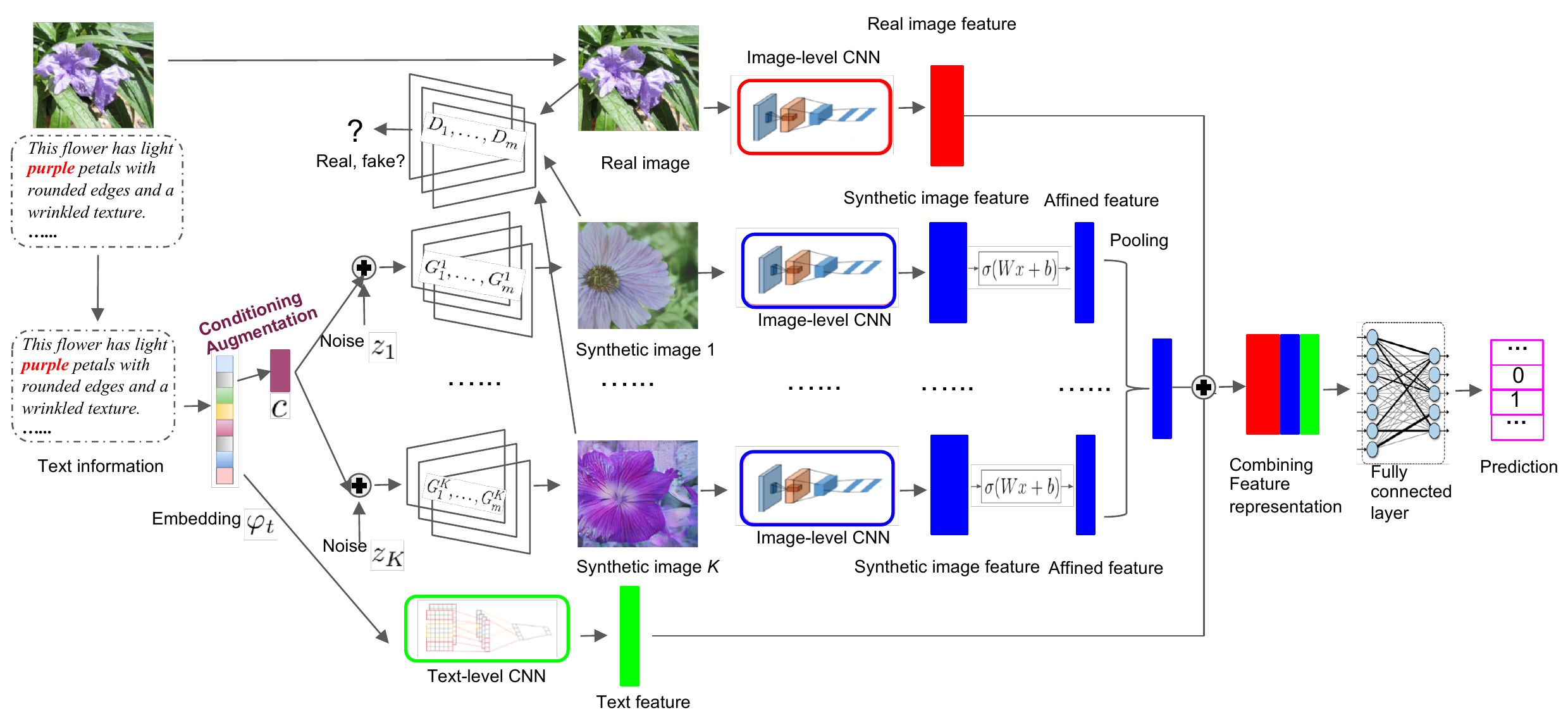}
\end{center}
\vspace{-0.40cm}
\caption{Overview of our proposed framework VITAL for image labeling with text information. At the beginning, we train a Text-to-image GAN, StackGMAN++, a modified version on StackGAN++~\cite{zhang:Arxiv2017}, to generate $K$ high-resolution and photo-realistic synthetic images.  Given an input image with its text information, we apply the trained StackGMAN++ to generate $K$ synthetic images. We apply an image-level CNN to extract visual feature on each synthetic image. To avoid the possible affine transforms between the synthetic images, we apply a linear transformation with ReLU to reduce the dimensions of the visual features. After that, we apply an element-wise pooling to extract visual feature representation in blue from all $K$ synthetic images. Finally, we combine the image-level feature (in red), text feature (in green) and the text corresponding visual feature (in blue) together by concatenation and feed them into a fully connected layer as a classifier to achieve the final prediction of labeling on the given image.}
\label{fig:gancnnpipeline}
\vspace{-0.30cm}
\end{figure*}

To sum up, our paper has three contributions:
\begin{itemize}
\item We propose a novel way to interpret text information with $K$ visually plausible synthetic images generated via our StackGMAN++ derived by modifying the number of generators at each stage from $1$ to $K$ on StackGAN++ associated with different noise priors.
\item We extract a common and compact visual feature representation from synthetic images conditioned on a text, and combine it with an image-level feature from real image, and a text-level feature together by conctatenatenation to boost the performance of image labeling. 
\item The experiments on two benchmark datasets have demonstrated that our proposed approach outperforms the state-of-art method~\cite{Johnson:ICCV2015}.
\end{itemize}

\section{Related work}
The related work can be divided into two categories: {\em text information for image labeling} and {\em Text-to-image GANs}.

{\bf Text information for image labeling}. As an important source for multimedia, text information has been studied and combined with image content to improve the accuracy of image labeling, because it provides informative and semantic signals~\cite{Yeh:CVPR2018} like who took the photo and, where and with whom. McAuley {\em et al.}~\cite{McAuleyECCV12} selected top-$1000$ most frequently occurring words from the texts in the training set as tags~\cite{guillaumin2010, Lindstaedt2008, sigurbjoernsson2008, SawantDLW10, Stone2008, Hu:CVPR2016learning}, explored pairwise social relations and applied a CRF-based structural learning approach for multi-label image annotation, which demonstrates impressive results although only metadata is used without any visual features. Following~\cite{McAuleyECCV12}, Johnson {\em et al.}~\cite{Johnson:ICCV2015} proposed a deep convolutional neural network to combine both the visual information of images and their neighbor images defined based on the shared similar metadata especially tags~\cite{Liu:CVPR2017semantic} from a text. With multiple text CNNs~\cite{Kim:EMNLP2014, Kalchbrenner:ACL2014, Johnson:NAACL15} emerged, Long {\em et al.}~\cite{Long:Arxiv2018} proposed to extract deep text-level features rather than user-generated tags for text representation in image labeling. It worths mentioning that Johnson {\em et al.}'s work~\cite{Johnson:ICCV2015} is a visual interpretation of text, although neighbor images depend too much on the density distribution of training data. Different from~\cite{Johnson:ICCV2015}, we propose to use a Text-to-image GAN to generate high-resolution and photo-realistic synthetic images to extract visual feature representation for the text of query image only, which we expect to be complementary to both visual feature on real images and textual feature on a text.

{\bf Text-to-image GANs} are proposed to effectively translate visual concepts from characters to pixels, with the aim to bridge these advances in text and image modeling~\cite{Plummer:ECCV2018, Liu:CVPR2018}. Based on text descriptions, Reed {\em et al.}~\cite{Reed:ICML2016} was able to generate plausible $64 \times 64$ images for birds and flowers, and $128 \times 128$ images were successfully generated by utilizing additional annotations on object part locations~\cite{Reed:NIPS2016}. To generate high-resolution ({\em e.g.}, $227 \times 227$) and photo-realistic images, Nguyen {\em et al.}~\cite{Nguyen:CVPR2017} proposed to generate images conditioned on a text using an approximate Langevin sampling approach with an iterative optimization process. Zhang {\em et al.} presented a Stacked GANs (StackGAN)~\cite{zhang:Arxiv2017stackgan} to generate $256 \times 256$ photo-realistic synthetic images with two stages, in which low-resolution images are generated at first stage, and then more details are added in the second stage to form high-resolution images with better quality by utilizing an encoder-decoder network before the upsampling layers. And its imporoved version StackGANN++~\cite{zhang:Arxiv2017} extends to multi-stage GANs with multiple generators and multiple discriminators arranged in a tree-like structure. Based on StackGAN and StackGAN++, Xu {\em et al.}~\cite{Xu:CVPR2018} adopted attention mechanism to generate synthetic images from fine-grained text with Attention Generative Adversarial Networks. We argue that high-resolution and photo-realistic synthetic images should be beneficial for us to extract visual representation for a text. Hence we modify the StackGAN++ to be a Generative Multi-Adverisal Network (GMAN), StackGMAN++, to generate $K$ synthetic images with different generators. Different from Durugkar {\em et al.}'s ~\cite{Durugkar:Arxiv2016} GMAN with one generator and multiple discriminators, our StackGMAN++ uses multiple generators with various prior noise vectors and one discriminator at each stage to generate multiple synthetic images, to well represent visual concepts embedded in text information.

\section{Proposed method}

As illustrated in Figure~\ref{fig:gancnnpipeline}, our proposed framework consists of three key components, {\em i.e.}, StackGMAN++ to generate $K$ synthetic images, extract visual feature representation from the synthetic images, and combine the synthetic image feature with real image feature and text feature to conduct a classification task for image labeling. We are going to discuss with details in the following subsections.

\subsection{StackGMAN++ to generate $K$ synthetic images}
Text-to-image Generative Adversarial Networks (GANs)~\cite{Reed:ICML2016, zhang:Arxiv2017stackgan, zhang:Arxiv2017} are proved to be able to generate visually-plausible images conditioned on a text. We start on StackGAN++~\cite{zhang:Arxiv2017} because it is able to generate high-resolution and
photo-realistic images. 
Note there are two implementation versions for StackGAN++, {\em i.e.}, StackGAN-v1 and StackGAN-v2. Our StackGMAN++ is derived from StackGAN-v2 which extends StackGAN-v1 from two stages to multiple stages and organizes generators and discriminators in a tree-like structure. 

Intuitively, given a text information, different people may imagine a different visual scene, and one text-based synthetic image is not sufficient to cover the underlying information behind the text itself. Therefore, to better explore the visual feature representation of any text information, we proposed to generate $K$ synthetic images for each text information. 

Different from StackGAN-v2 which takes only one noise vector $z$ as the input and has a generator $G_i$ and a corresponding discriminator $D_i$ for $i$-th branch with different scales, our StackGMAN++ takes a conditional vector $c$ which is defined based on the text embedding $\varphi_t$ in~\cite{zhang:Arxiv2017} and $K$ different prior noise vectors $z_1, \ldots, z_K$ (we smaple values for each $z_k$ from a normal distribution) as input, and at each stage $i$, we design one discriminator $D_i$ and $K$ generators  $G_i^1, \ldots, G_i^K$ to generate synthetic images at a certain scale. We pass the hidden feature $h_i^k$ for each generator $G_i^k$ by a non-linear transformation,
\begin{equation}
h_i^k = \left\{\begin{matrix}
 & F_i^k(c, z_k) ~~~~~~~~~~~~~~~~~~~~~ i = 0\\ 
 & F_i^k(h_{i-1}^k, c) ~~~~~ i = 1, \ldots, m
\end{matrix}\right.
\end{equation}
where $h_i^k$ represents hidden features for the $k$-th generator $G_i^k$ at the $i$-th branch, $m$ is the total number of branches, and $F_i^k$ is modeled as the corresponding neural network. In order to encourage the generators to draw images with more details according to the conditioning variables, $c$ is concatenated to the hidden features $h_{i-1}^k$ as the inputs of $F_i^k$ for calculating $h_i^k$. Based on hidden features at different layers $(h_1^k, \ldots, h_m^k)$, generators $G_1^k, \ldots, G_m^k$ can generate synthetic images of small-to-large scales
\begin{equation}
s_i^k = G_i^k(h_i^k), i = 1, \ldots, m.
\end{equation}

For the training purpose, we define the loss functions with joint conditional and unconditional distribution approximation at each stage $i$ as following:
\begin{equation}
\begin{split}
\mathcal{L}_{D_i} &= K \mathbb{E}_{{\bf x}_i \sim p_{{data}_i}} [\text{log} D_i({\bf x}_i)] + \\
& \sum\limits_{k=1}^K \mathbb{E}_{s_i^k \sim p_{G_i^k}} [\text{log}(1 - D_i(s_i^k)] + \\
& + K \mathbb{E}_{{\bf x}_i \sim p_{{data}_i}} [\text{log} D_i({\bf x}_i, c)] + \\
& \sum\limits_{k=1}^K \mathbb{E}_{s_i^k \sim p_{G_i^k}} [\text{log}(1 - D_i(s_i^k, c)], \label{eqn:lossDi}
\end{split}
\end{equation}
\begin{equation}
\begin{split}
\mathcal{L}_{G_i^1, \ldots, G_i^K} &= \sum\limits_{k=1}^K \mathbb{E}_{s_i^k \sim p_{G_i^k}} [-\text{log} D_i(s_i^k)] \\
&+ \sum\limits_{k=1}^K \mathbb{E}_{s_i^k \sim p_{G_i^k}} [-\text{log} D_i(s_i^k, c)] \label{eqn:lossGi}
\end{split}
\end{equation}
where ${\bf x}_i$ is from the true image distribution $p_{{data}_i}$ at the $i$-th scale, and $s_i^k$ is from the model distribution $p_{G_i^k}$ at the same scale. Then the discriminator $D_i$ and generators $G_i^1, \ldots, G_i^K$ at $i$-th stage can be optimized in a joint form by alternatively maximizing $\mathcal{L}_{D_i}$ and minimizing $\mathcal{L}_{G_i^1, \ldots, G_i^K}$ until convergence.

Based on StackGMAN++, 
we are able to generate $K$ high-resolution and photo-realistic synthetic images ${\bf s} = \{s^1, \ldots, s^K\}$ with $s^k = G^k(s_{m-1}^k)$, and the size of each image is $256 \times 256$.



{\bf Discussion}: we extend StackGAN++ to StackGMAN++ and joint learn the model with a united loss function to generate the diverse synthetic images. Note that the $K$ generators in StackGMAN++ share weights. In this way, we can control the training cost when compared to train $K$ StackGAN++ separately. We observe that training a StackGMANN++ with shared weights is less expensive than training $K$ StackGAN++, and the $K$ synthetic images generated by StackGMANN++ are more diverse. 

\subsection{Visual representation for text Information}
Given a text $t$, we are able to generate $K$ synthetic images ${\bf s}=\{s^1, \ldots, s^K\}$ with our StackGMAN++. Then we feed $K$ generated synthetic images into ResNet~\cite{HeResnet:CVPR2016} to extract visual feature. Note that we use ResNet as feature extractor $\phi$ in the pretrained ResNet model~\cite{HeResnet:CVPR2016}, and get a feature map of the second last layer of $\phi(s^k)$ of $7 \times 7 \times 2048$ size for each synthetic image $s^k$, respectively. 

\begin{figure}
\begin{center}
\includegraphics[width=0.95\linewidth, angle=0]{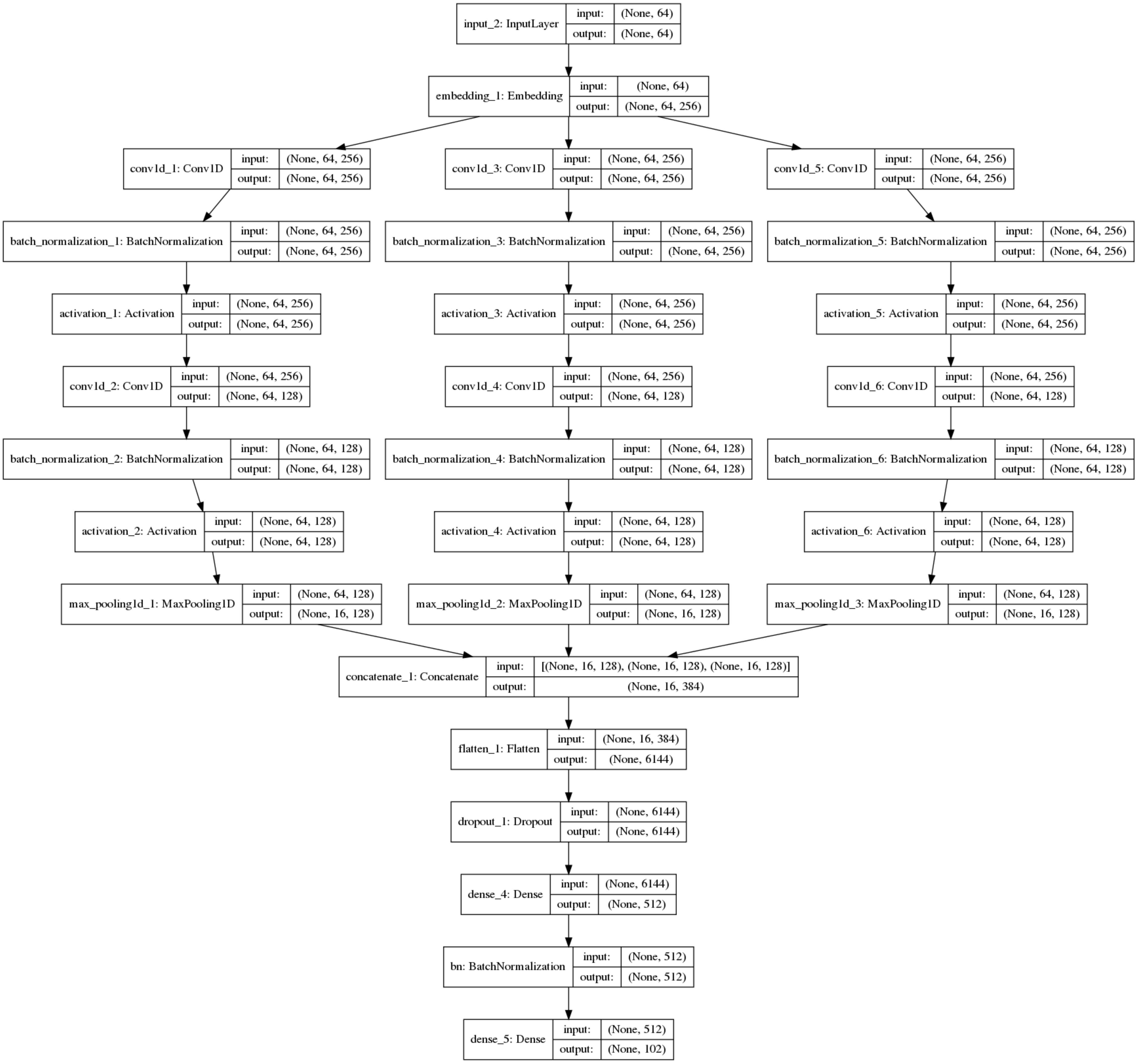}
\end{center}
\vspace{-0.60cm}
\caption{Plot of the multichannel convolutional neural network, 3Text-CNNs, for text-level feature extraction.}\label{fig:3textcnns}
\vspace{-0.30cm}
\end{figure}

In order to fuse visual features for these $K$ synthetic image, we compute an $h$-dimensional hidden state for each image by applying an affine transformation and an element-wise ReLU nonlinearity $\sigma(\varepsilon) = \text{max}(0, \varepsilon)$ to its feature. To let the model treat hidden states for each synthetic image differently, we apply distinct transformations to $\phi(s^k)$ with parameters $W_k \in \mathbb{R}^{d\times h}$ and $b_k \in \mathbb{R}^h$, and then we arrives at hidden states ${\bf v}_{s^k} \in \mathbb{R}^h$ for $s^k \in {\bf s}$. To generate as single hidden state ${\bf v}_s \in \mathbb{R}^h$ for all the synthetic images ${\bf s}$, we apply an element-wise max-pooling on each ${\bf v}_{s^k}$ so that ${\bf v}_s = \text{max}_k {\bf v}_{s^k}$, {\em i.e.}, 
\begin{equation}
{\bf v}_s = \text{max}_k(\sigma(W_k\phi(s^k) + b_k))
\end{equation}
and pass it to be included into the final combined feature representation with image feature and text feature. 


{\bf Discussion}: we choose synthetic images rather than visual features because we want to visually interpret the text with high quality synthetic images so that the human can view directly, while visual features may miss some details.

\subsection{Feature fusion for image labeling}\label{sec:featurefusion}

As shown in Figure~\ref{fig:gancnnpipeline}, besides the visual feature representation ${\bf v}_s$ on $K$ synthetic images, we also include the text-level feature ${\bf v}_t$ and image-level feature ${\bf v}_x$ together to better explore both image and text information to improve the quality for image labeling. In this paper, we choose to use the second last layer of ResNet~\cite{HeResnet:CVPR2016} to extract the visual feature ${\bf v}_x$ for real image ${\bf x}$ because ResNet has been proved successful in most visual application tasks. 

Regarding text information, a standard deep learning model for text classification and sentiment analysis uses a word embedding layer and one-dimensional convolutional neural network~\cite{Kim:EMNLP2014}. The model can be expanded by using multiple parallel convolutional neural networks that read the source document using different kernel sizes. This, in effect, creates a multichannel convolutional neural network for text that reads text using different $n$-gram sizes (groups of words). We follow Kim's multi-channel model to implement a merged model with 3 text CNNs with kernels of different sizes, denoted as 3Text-CNNs, as illustrated in Figure~\ref{fig:3textcnns},
to extract 512-dimensional text-level feature ${\bf v}_t$ from the second last layer. 

All these three kinds of features are combined to form the final feature representation by concatenation as $({\bf v}_x, {\bf v}_{\hat x}, {\bf v}_t)$ and feed them into a fully connected layer to conduct a classification task as image labeling.

\subsection{Implementation details}

The parameters need to be learned include the parameters in $mK$ generators and $m$ discriminators in StackGMAN++, parameters in ResNet-50s, 3Text-CNNs, the affine transformation parameter $W_{\hat x}$ and $b_{\hat x}$, and the parameters in the last fully connected layer. Note our training procedure is divided into two phases. At Phase I, we use pairs of text and its corresponding image to train our StackGMAN++ in an alternative optimization procedure until it convergence. Then at Phase II, we apply the trained StackGMAN++ to generate $K$ high-resolution and photo-realistic synthetic images. Then we feed $n$ real images, $nK$ synthetic images and $n$ text to learn the rest of parameters in the entire framework with cross-entropy loss function. 

Note that we follow the tricks in StackGAN-v2 to train StackGMAN++ at each stage at Phase I with a batch size of 12 for
350000 iterations. At Phase II, with a minibatch size of 50, we initialize all parameters with pre-trained models (ResNet and 3Text-CNNs) and use stochastic gradient descent with a fixed learning rate, RMSProp, as the optimization.

\section{Experiments}

\begin{figure*}
\begin{center}
\vspace{-0.60cm}
\subfloat[VITAL-S]{{\includegraphics[width=.32\linewidth]{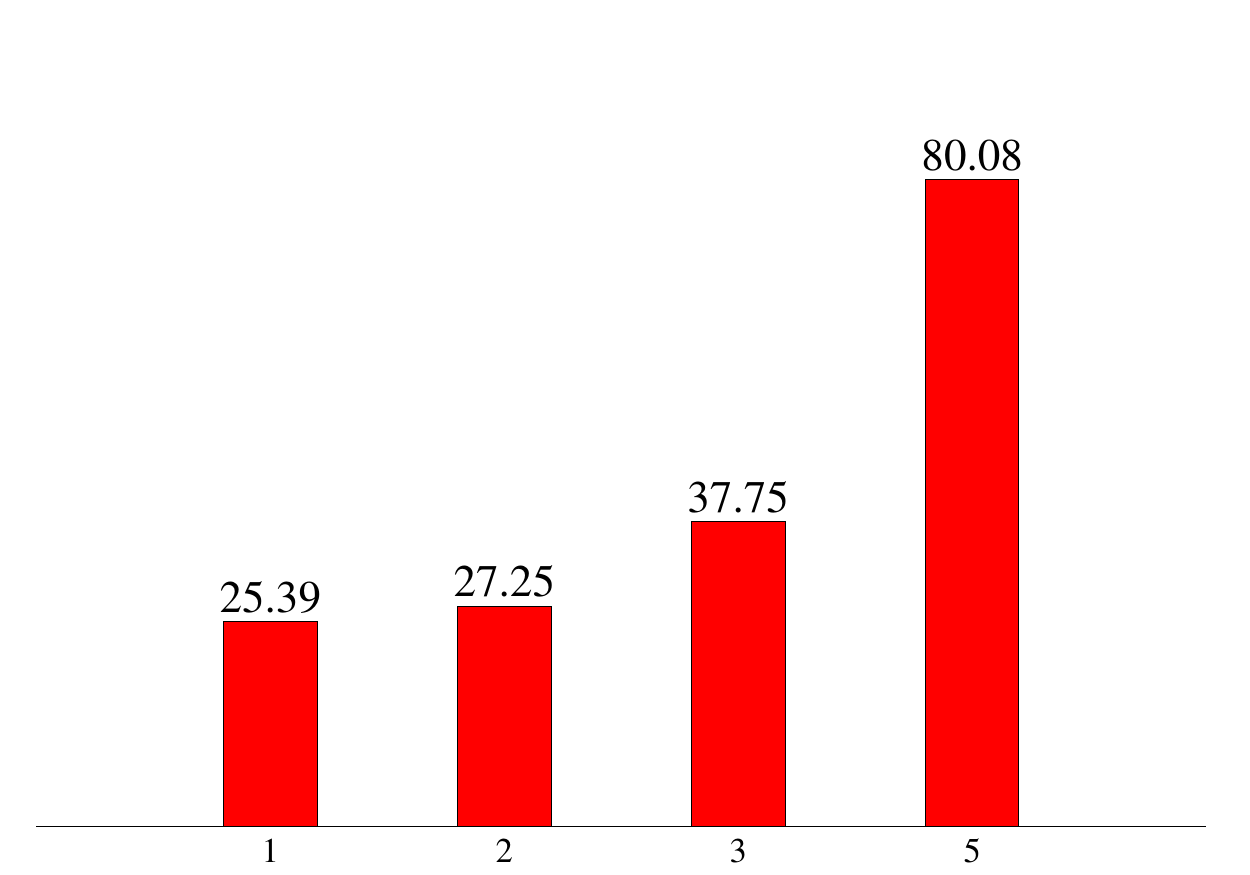} }} \label{fig:analyze_follower_haming_dist}
\subfloat[VITAL-RS]{{\includegraphics[width=.32\linewidth]{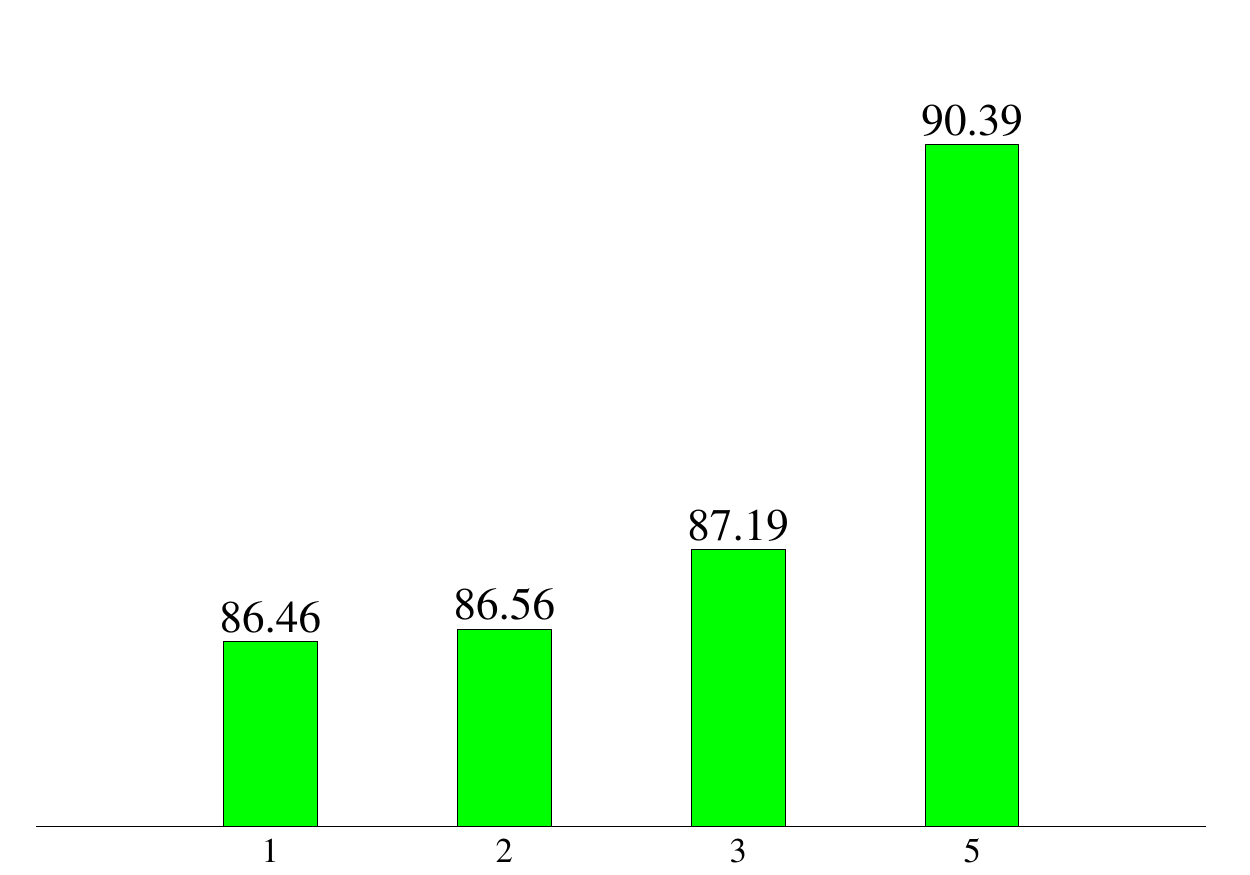} }} \label{fig:analyze_follower_cosine_dist}
\subfloat[VITAL-RST]{{\includegraphics[width=.32\linewidth]{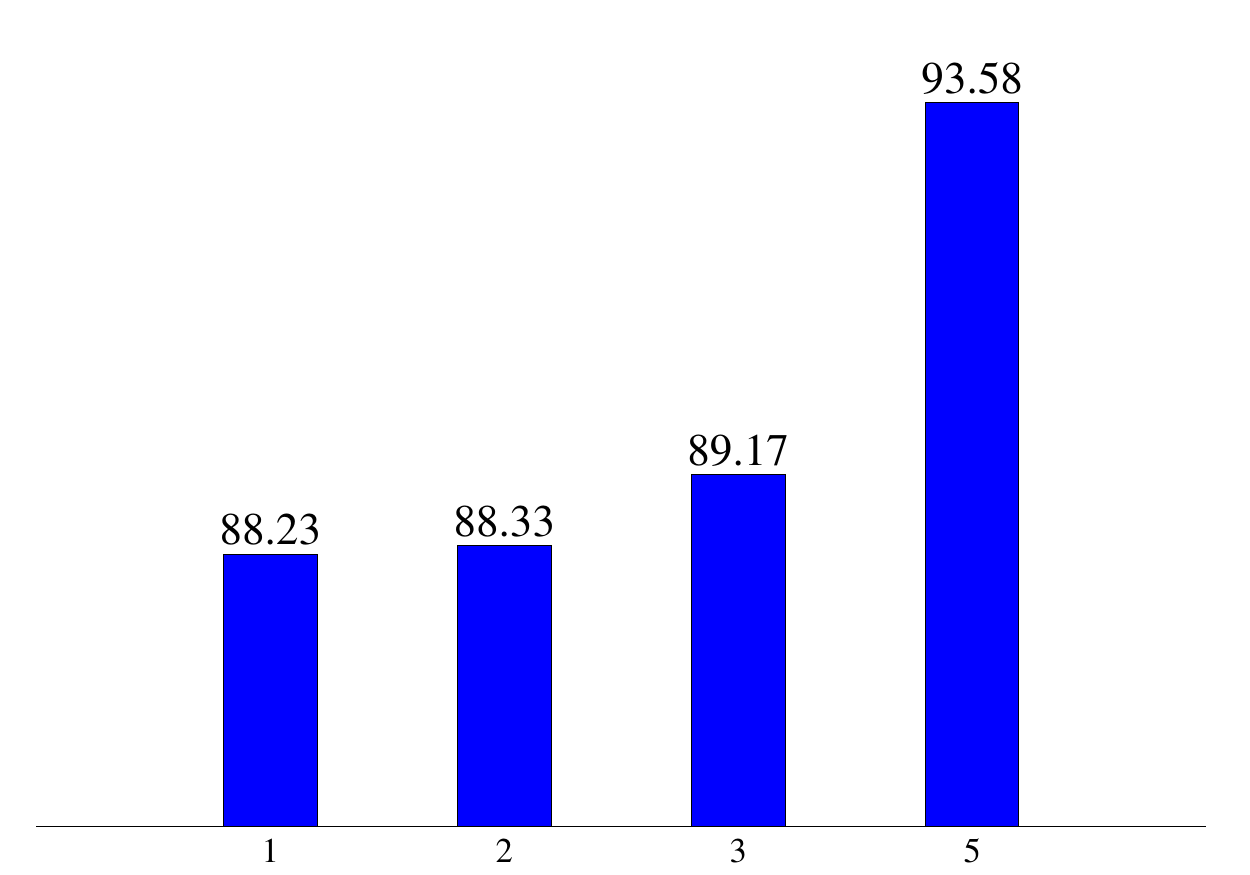} }} \label{fig:analyze_follower_cosine_dist}
\end{center}
\vspace{-0.60cm}
\caption{Performance with accuracy (unit: \%) for our VITAL-S, VITAL-RS and VITAL-RST using visual feature representation on $K$ synthetic images with different values of $K$, {\em i.e.}, $K = 1, 2, 3,$ and $5$ on the 102 Category Flower Dataset.}\label{fig:visualfeature_synthetic_Kvalue}
\vspace{-0.30cm}
\end{figure*}

Our experiments are conducted on two datasets, {\em i.e.}, the Oxford 102 Category Flower Dataset~\cite{Nilsback:2008automated},  and the Caltech-UCSD Birds-200-2011 Dataset~\cite{Wah:CUB_200_2011}. 
We use accuracy as the metric to measure performance of image labeling.

\subsection{Experiments on the  Oxford 102 Category Flower Dataset}

The Oxford 102 Category Flower Dataset~\cite{Nilsback:2008automated} consists of 8,189 images from 102 categories of flowers which commonly occurs in the United Kingdom, and each category has 40 to 258 images. The images have large scales, pose and light variations. The text context information is provided by~\cite{Reed:CVPR2016learning} with 10 descriptions for each image. Due to the limit space, we plot text using only 1 sentence and use the symbol ``{\bf ......}" to represent the rest 9 sentences in this paper. We train our StackGMAN++ with text and the corresponding real images. With the learned StackGMAN++, we are able to generate $K$ visual plausible synthetic images conditioned on a text for experiments. 

\begin{figure}
\begin{center}
\includegraphics[height=0.55\linewidth, width=1.01\linewidth, angle=0]{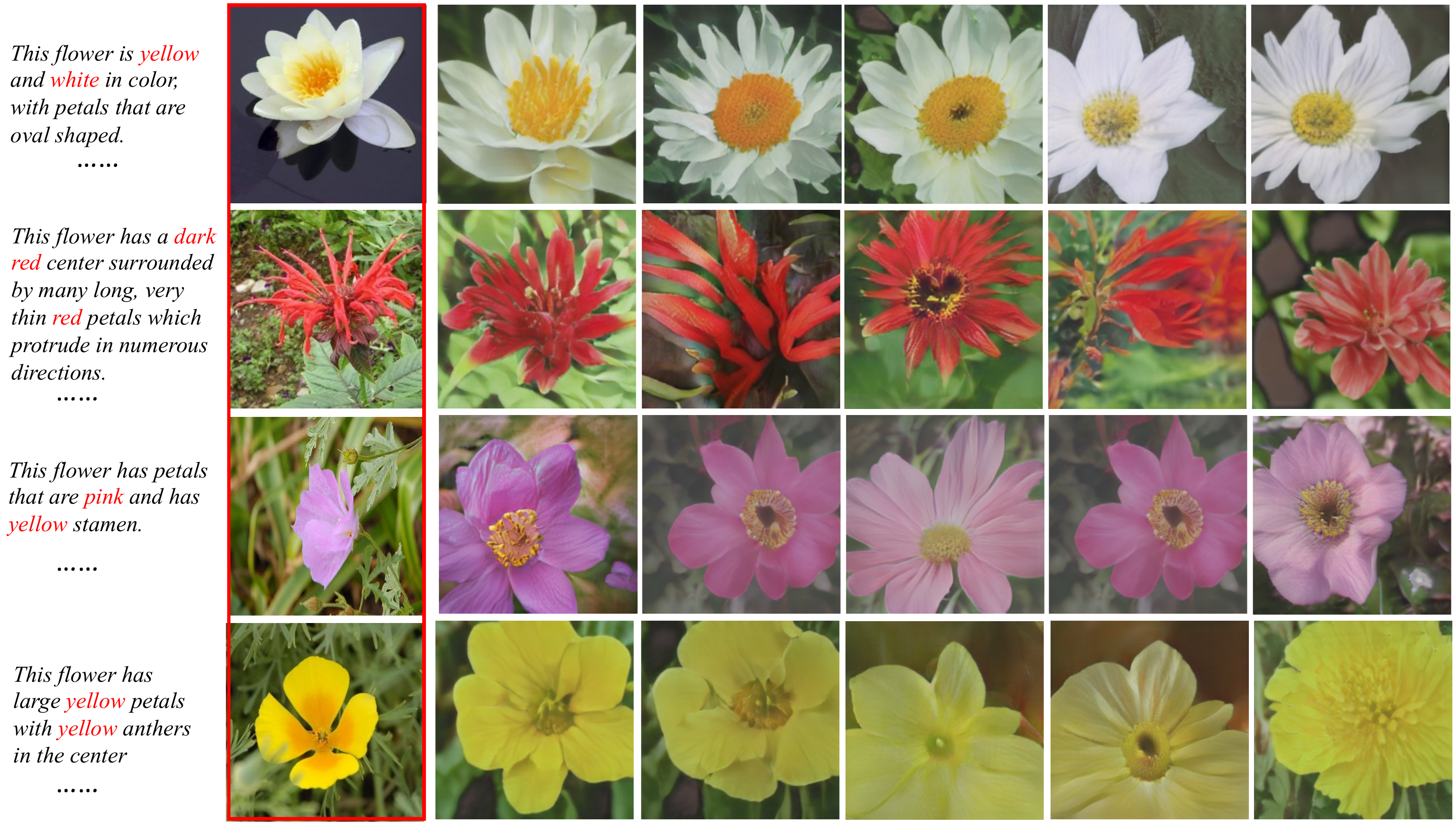}
\end{center}
\vspace{-0.40cm}
\caption{Visualization of $K=5$ high-resolution and photo-realistic synthtic images (blue) condtioned on a text, and compared with the corresponding real images (red) on the Oxford 102 Category Flower Dataset.}
\label{fig:visualexample_flowers}
\end{figure}

\subsubsection{Effectiveness of StackGMAN++}

To verify the effectiveness of our StackGMAN++, we conduct experiments to check visual concept consistency between the generated synthetic images by StackGMAN++ and the corresponding real images, and the effectiveness of visual feature representation on $K$ synthetic images. 

{\bf Visual concept consistency between synthetic and real images}. We firstly visualize the generated synthetic images conditioned on a text and measure the correlation between our generated synthetic images with the corresponding images in a visual feature space.

As shown in Figure~\ref{fig:visualexample_flowers}, our generated $K$ synthetic images are able to not only cover main content elements in the text, but also provide underlying background and other rich visual information like size, shape and pose variations which are not mentioned in the text, due to various prior noise vector $z_k$. These observations are consistent with our human's behavior to interpret a text based on his/her prior knowledge. In other words, even given the same text, different people with different growing backgrounds will imagine different visual pictures in their brains. Such diverse interpretations are complementary to each other and can be merged to formulate a more representative format.

\begin{figure}
\begin{center}
\subfloat[With Hamming distance.]{{\includegraphics[width=.48\linewidth]{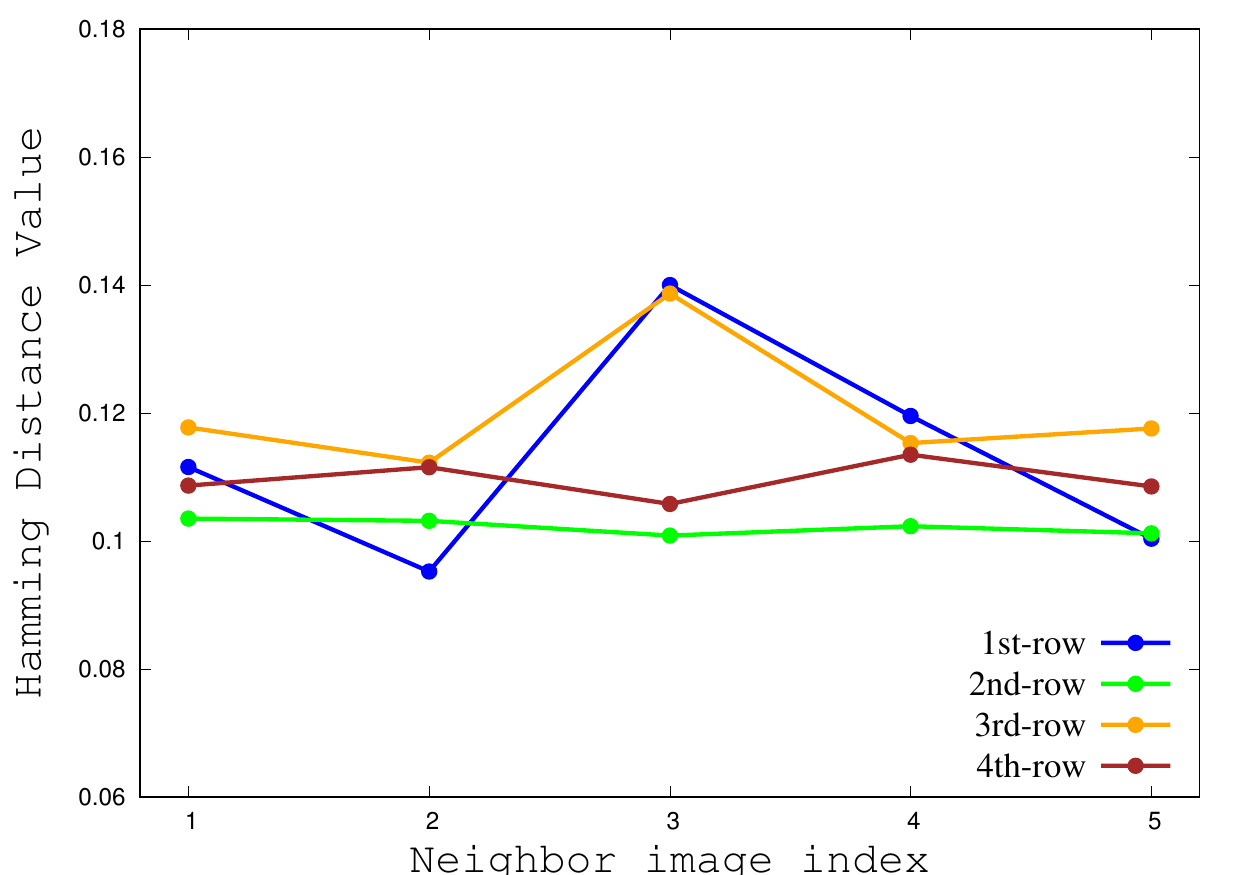} }} \label{fig:follower_haming_dist}
\subfloat[With Cosine similarity.]{{\includegraphics[width=.48\linewidth]{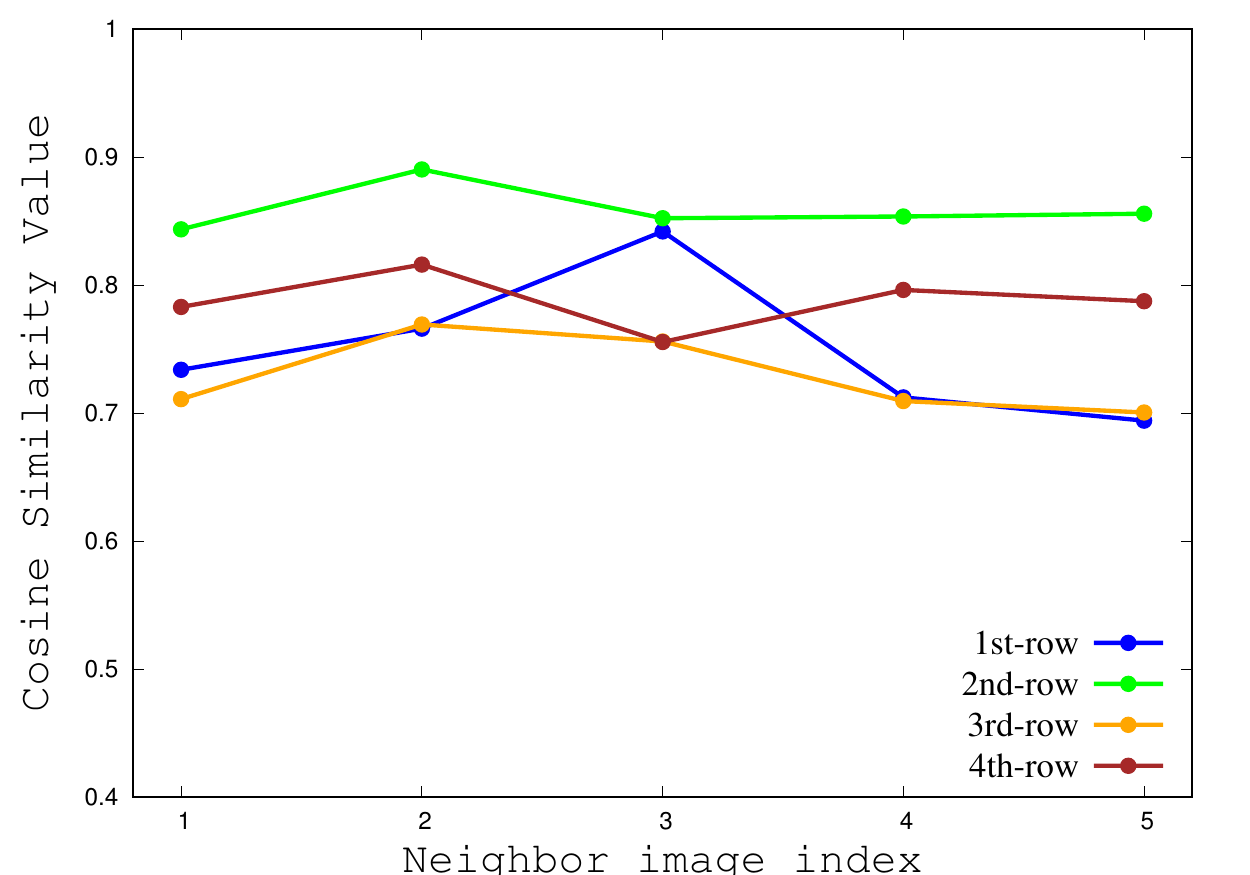} }} \label{fig:follower_cosine_dist}
\end{center}
\vspace{-0.60cm}
\caption{The correlation measured with Hamming distance and Cosine similarity between the synthetic images generated by our StackGMAN++ conditioned on a text and the corresponding real images on the  Oxford 102 Category Flower Dataset.} \label{fig:similarityFlower}
\vspace{-0.30cm}
\end{figure}

We also measure the correlation between our generated synthetic images conditioned on a text with the corresponding real images with two distance metrics, {\em i.e.}, Hamming distance and Cosine similarity, between their visual feature vectors extracted by ResNet~\cite{HeResnet:CVPR2016}. For Hamming distance in the range $[0, 1]$, smaller value means the higher similarity between images.  
For Cosine similarity in the range $[-1, 1]$, the value closer to 1.0 means two compared images are more correlative to each other in the given feature space. We plot both Hamming distance values and Cosine similarity value between our generated synthetic images and the corresponding real images in Figure~\ref{fig:similarityFlower}. As we observe, none of Hamming distances is larger than 0.15 and all Cosine distance values are over 0.70, which indicates high correlation between each synthetic image and the corresponding real image in the visual feature space.


{\bf Effectiveness of visual feature representation on $K$ synthetic images}. 
As stated in Section~\ref{sec:featurefusion}, our VITAL uses the feature combinations $({\bf v}_x, {\bf v}_s, {\bf v}_t)$ where ${\bf v}_x$ and ${\bf v}_s$ indicate the visual feature representation extracted by ResNet~\cite{HeResnet:CVPR2016} on real image, and synthetic images, respectively, and ${\bf v}_t$ represents the text-level feature extracted from 3Text-CNNs~\cite{Yin:Arxiv2016multichannel}. We then develop two different baseline algorithms with two different feature combinations, {\em i.e.}, ${\bf v}_s$ only and $({\bf v}_x, {\bf v}_s)$. 
For notation simplification, we denote our proposed method to be VITAL-RST where ``-RST" represents the feature combinations $({\bf v}_x, {\bf v}_s, {\bf v}_t)$. We further denote the first to second baseline algorithm to be VITAL-S and VITAL-RS, respectively. 

We conduct a group of experiments by setting $K$ to be various values, {\em i.e.}, 1, 2, 3 and 5. The results are summarized in the Figure~\ref{fig:visualfeature_synthetic_Kvalue}. Apparently, for any algorithm among VITAL-S, VITAL-RS, and VITAL-RST, the performance accuracy goes up when the value of $K$ increases. This indicates that multiple synthetic images are complementary to be used for extracting visual concepts embedded in text and boosting the accuracy for image labeling.  

Note that $K$ is also the number of generators in each stage in our StackGMAN++. Therefore, when $K=1$, our StackGMAN++ is equivalent to StackGAN++ with its implementation version StackGAN-v2. Obviously, all these three algorithms with $K$ (especially when $K > 1$) synthetic images generated by our StackGMAN++ always work better than using only synthetic images generated by StackGANN++. The observation strongly demonstrates the effectiveness and robustness of our StackGMAN++.

\subsubsection{Performance comparison}

\begin{table}[]
\centering
\caption{Performance comparison on the Oxford 102 Category Flower Dataset.  (unit: \%)}
\vspace{-0.20cm}
\begin{tabular}{c|c}
\hline
Methods & Accuracy \\ \hline
\hline
JCNN-NN~\cite{Johnson:ICCV2015} & 89.16$\pm$0.57 \\ \hline
3Text-CNNs~\cite{Kim:EMNLP2014} & 37.63$\pm$0.76 \\ \hline
ResNet~\cite{HeResnet:CVPR2016} & 85.86$\pm$1.85 \\ \hline
ResNet~\cite{HeResnet:CVPR2016}+3Text-CNNs~\cite{Kim:EMNLP2014} & 88.39$\pm$0.44 \\ \hline
VITAL-S  & 79.87$\pm$0.31 \\ \hline
VITAL-RS & 89.68$\pm$0.61 \\ \hline
VITAL-RST & {\bf 93.38}$\pm$0.15 \\ \hline
\end{tabular}\label{table:performance_comparison_flowers}
\vspace{-1.50cm}
\end{table}

\begin{figure}
\begin{center}
\includegraphics[width=0.88\linewidth, angle=0]{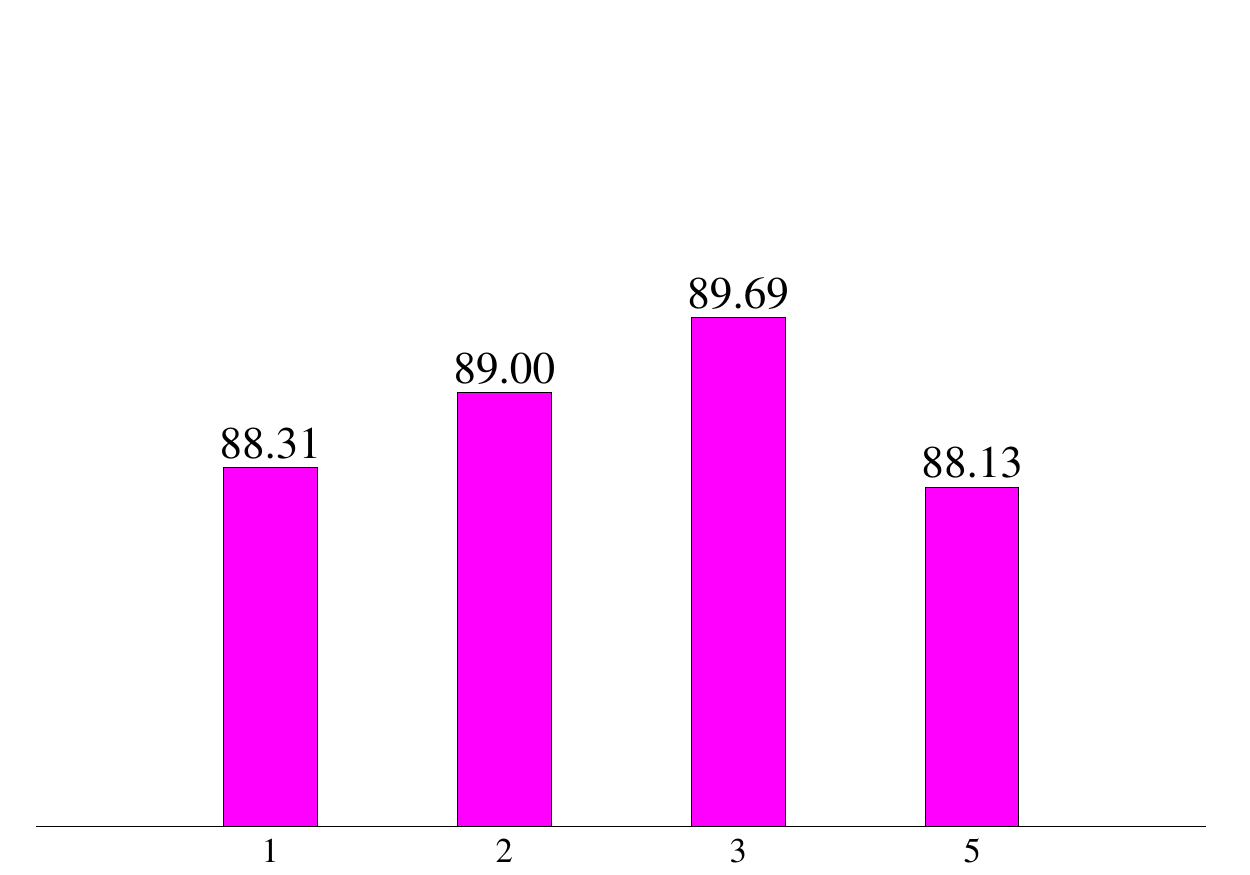}
\end{center}
\vspace{-0.60cm}
\caption{Performance with accuracy (unit: \%) for JCNN-NN with different $K$ nearest neighbor images. }\label{fig:JohnCNNwNN_K_Values}
\vspace{-0.30cm}
\end{figure}

We compare our proposed VITAL-RST with Johnson {\em et al.}'s Convolutional Neural Network with Nearest Neighborhood~\cite{Johnson:ICCV2015}, denoted as ``JCNN-NN", which explores the related neighbor images to interpret image metadata especially including tags from text information. To our best acknowledge, JCNN-NN is the most closely related work to our VITAL because it can be interpreted as a visual interpretation of image metadata with the related neighbor images. To make sure the comparison is fair, we utilize the same ResNet~\cite{HeResnet:CVPR2016} as the visual feature extractor in JCNN-NN. In addition, we add two simple baseline algorithms, {\em i.e.}, ResNet~\cite{HeResnet:CVPR2016} and 3Text-CNNs~\cite{Yin:Arxiv2016multichannel}, which indicate using ${\bf v}_x$ only and using ${\bf v}_t$ only, respectively.

To clarify, we do not compare our VITAL with StackGMAN++ to other adversarial learning methods because this is not our focus and our focus is how to interpret text for image labeling by extending and applying the existing text-to-image GANs. We run experiments on the Oxford 102 Category Flower Dataset. Note that we also run JCNN-NN with five different number of neighbor images, as shown in Figure~\ref{fig:JohnCNNwNN_K_Values}, from which we can find $K=3$ is the best choice for JCNN-NN.  

We conduct the experiments repeatedly for 10 times with 10 different random data split to form the training and testing sets and the results with all six algorithms are summarized in Table~\ref{table:performance_comparison_flowers}, from which we can observe: (1) 3Text-CNNs performs much worse than ResNet by a half and this conveys a clue that text on the dataset is a little weaker when compared with image content; (2) with single feature representation, our VITAL-S works much better than 3Text-CNNs and close to ResNet, 
which indicates visual feature extracted on synthetic images is more representative then text-feature; (3) combining with real image feature, VITAL-RS is able to improve the performance by 4.01\% from ResNet, and works better than JCNN-NN~\cite{Johnson:ICCV2015} and ResNet~\cite{HeResnet:CVPR2016}+3Text-CNNs~\cite{Kim:EMNLP2014} ({\em i.e.}, a combination of RT without involving VITAL), which shows that our visual interpretation on text is more effective and robust than using a set of neighbor images defined based on the Jaccard similarity between image metadata especially tags extracted from text; 
and (4) combining with both real image feature and text feature, VITAL-RST performs the best. Apparently, the visual interpretation in our VITAL is robust and the visual synthetic image feature is complementary to both visual real image feature and text feature. 

%
%


\subsection{Experiments on the Caltech-UCSD Birds-200-2011 Dataset}

The Caltech-UCSD Birds-200-2011 Dataset consists of 11,169 
bird images from 200 categories and each each category has 60 images averagely. We randomly select 9,935 images for training, and use the resting 1,234 images for testing. The dataset is very chanllenging because it contains images with multiple objects and various backgrounds. We train our StackGMAN++ with $K=5$ and use it to generate synthetic images on the text with 10 descriptions for each real image to conduct the experimental evaluation.






We repeat the experiments with 10 different random training/testing data splitting and summarize the results in Table~\ref{table:newperformance_comparison_birds}, which can be observed from four aspects.  (1) ResNet performs much better than 3Text-CNNs, which indicates image is more representative than text content. (3) Using text information only, our VITAL-S still outperforms 3Text-CNNs. (3) VITAL-RST performs a little better than VITAL-RS which works much better than VITAL-S. Again, this observation confirms the complementary relationship between three kinds of feature representations. (4) Our VITAL-RS and VITAL-RST, performs better than JCNN-NN and ResNet+3Text-CNNs, which suggests that our proposed VITAL is good at visual interpretation on a text by extracting visual concepts from the text for boosting the labeling accuracy.

\begin{table}[]
\centering
\caption{Performance comparison on the resized Caltech-UCSD Birds-200-2011 dataset.  (unit: \%)}
\vspace{-0.20cm}
\begin{tabular}{c|c}
\hline
Methods & Accuracy \\ \hline
\hline
JCNN-NN~\cite{Johnson:ICCV2015} & 89.91$\pm$0.48 \\ \hline
3Text-CNNs~\cite{Kim:EMNLP2014} & 5.86$\pm$0.69 \\ \hline
ResNet~\cite{HeResnet:CVPR2016} & 86.81$\pm$1.14 \\ \hline
ResNet~\cite{HeResnet:CVPR2016}+3Text-CNNs~\cite{Kim:EMNLP2014} & 89.38$\pm$1.17 \\ \hline
VITAL-S  & 55.05$\pm$0.20 \\ \hline
VITAL-RS & 93.28$\pm$1.25\\ \hline
VITAL-RST & {\bf 94.49}$\pm${\bf 0.90} \\ \hline
\end{tabular}\label{table:newperformance_comparison_birds}
\vspace{-0.20cm}
\end{table}

\subsection{VITAL performance with GMAN vs. $K$ GANs}
Since we can train $K$ GANs with $K$ different noise levels seperately and then use the trained GANs to generate $K$ synthetic images, we also can assess the performance of our VITAL with $K$ GANs. For the purpose of fair comparison, we train $K=5$ StackGAN++ with different noise levels individually and use the $K$ generated synthetic images to extract visual features for image labeling on the Oxford 102 Category Flower Dataset. The accuracy performance of VITAL-RST with $K$ StackGAN++ is 90.59\%, which is 3\% lower than VITAL-RST with our StackGMAN++. 

\begin{figure}
\begin{center}
\includegraphics[width=0.90\linewidth, angle=0]{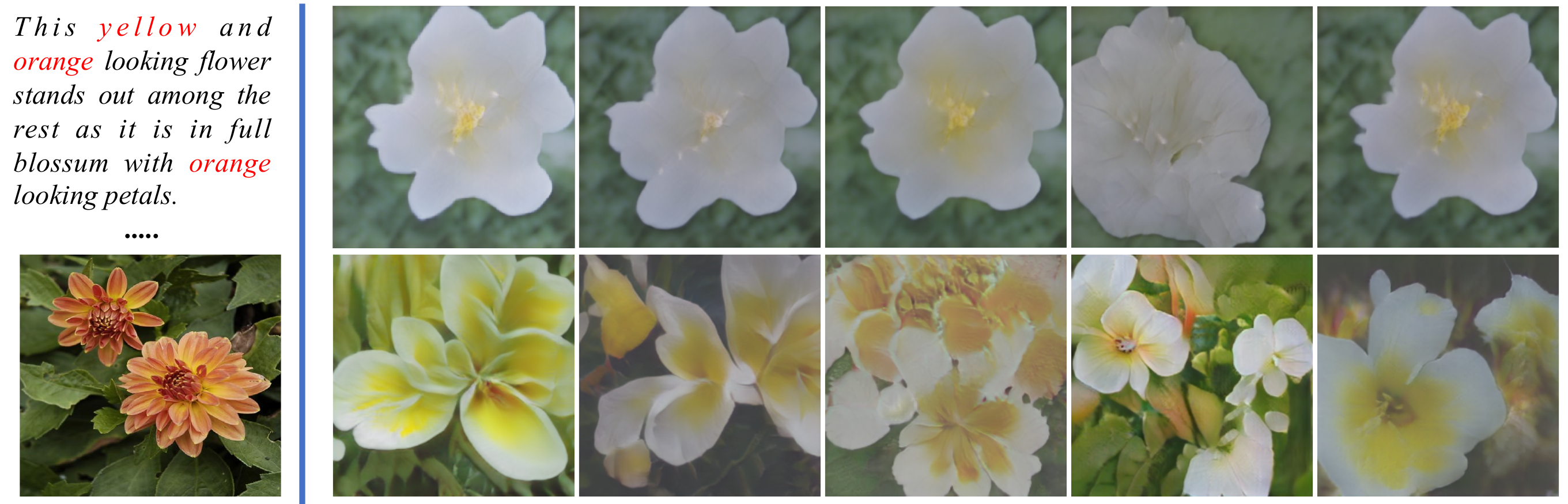}
\end{center}
\vspace{-0.60cm}
\caption{Visualization of $K=5$ syththtic images (blue) generated by StackMGAN++ (right bottom) and $K$ StackGAN++ (right top). The input text and the corresponding real image are on the left. }\label{fig:GMANvsKGANs}
\vspace{-0.30cm}
\end{figure}

We visualize the synthetic images generated by both StackGMAN++ and $K$ StackGAN++ in Figure~\ref{fig:GMANvsKGANs}. Obviously, our proposed StackGMAN++ is able to generate more diverse images, and the diversity among the synthetic images benefits the performance of image labeling.

\subsection{Analysis on successful and failure cases}

\begin{figure}
\begin{center}
\includegraphics[width=0.92\linewidth, angle=0]{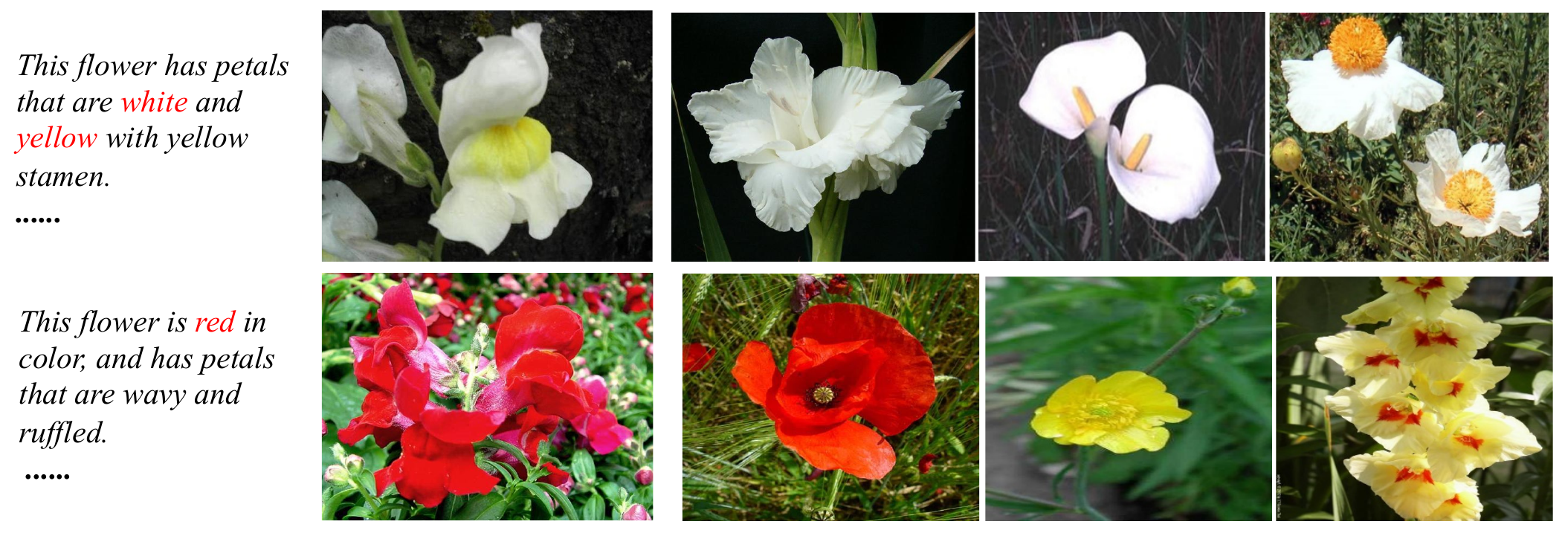}
\end{center}
\vspace{-0.60cm}
\caption{Visualization of JCNN-NN~\cite{Johnson:ICCV2015}'s $K=3$ nearest neighbor (NN) images based on the Jaccard 
similarity of tags extracted from text. From left to right are: text, real image, and 3 nearest-neigbour images, respectively.}\label{fig:visualization_CNNwNN}
\vspace{-0.30cm}
\end{figure}


To further explain why our proposed VITAL works better than JCNN-NN, we conduct an analysis on the quality of the nearest neighbor (NN) images used in JCNN-NN. As shown in Figure~\ref{fig:visualization_CNNwNN}, the color of neighbor images are not always consistent with real images, and the background of all the neighbor images are more complicated when compared with the generated synthetic images in Figure~\ref{fig:visualexample_flowers}. Moreover, the Jarccard similarity between the query image and the $k$-th NNs always decreases when the value of $k$ increases. 

As a non-parameter method, the quality of nearest neighbor images in JCNN-NN will be perfect if an infinite number of samples is available, but in practice the number of samples is limited. Therefore the quality of neighbor images depends on the density distribution in training data. If the distribution is dense enough, then the quality of neighbor images can be a guarantee. Obviously, in both the 102 Category Flower Dataset and Caltech-UCSD Birds-200-2011 Dataset, the observations indicate that the density is not sufficient to ensure the good quality of neighbor images for JCNN-NN.

On the contrary, our proposed VITAL generates $K$ visual plausible synthetic images by StackGMAN++ conditioned on a text of query image only, without requiring text for any other images, which makes the generated synthetic images are closely correlated to the corresponding real images, displaying key visual elements embedded in text and even providing more details about the underlying background and variations. Therefore, our proposed VITAL is more robust to visually interpret text for improving the performance of image labeling.

\begin{figure}
\begin{center}
\includegraphics[width=0.99\linewidth, angle=0]{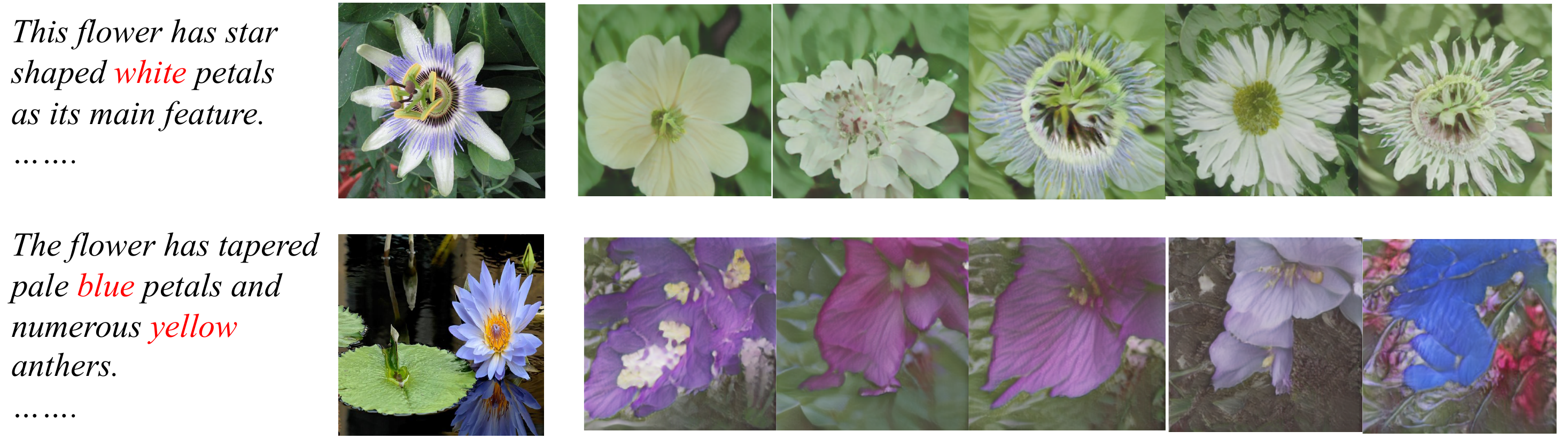}
\end{center}
\vspace{-0.60cm}
\caption{Visualization of a successful case (top) and a failure case (bottom) for our proposed VITAL. From left to right are: text, real image, and 5 synthetic images, respectively.}\label{fig:visualization_successfailurecase}
\vspace{-0.30cm}
\end{figure}

We also visualize a successful case and a failure case in Figure~\ref{fig:visualization_successfailurecase}. As we can observe, if text describes a flower with detailed information about colors hand shapes, then our StackGMAN++ is able to generate a bunch of informative synthetic images for leveraging the performance of image labeling. Otherwise, if a text is hard to understand, ambiguous and even misleading, then the generated synthetic images will not be consistent with each other to represent the visual concepts well.     

\section{Conclusion}

In this paper, we propose a novel way to visually interpret text with adversarial learning for image labeling. With our StackGMAN++, $K$ high-resolution and photo-realistic synthetic images are generated conditioned on a text to represent the visual concepts. The visual synthetic image feature has been proved to be able to improve accuracy for image labeling, and it is complementary to real image feature and text feature. The experimental results conducted on two datasets well support our claims in the paper.

Our future work includes further exploring our GMAN to incorporate the state-of-the-art GANs like AttnGAN~\cite{Xu:CVPR2018} to produce better quality of synthetic images, improving VITAL's performance on image labeling, and extending our VITAL to solve more general multimedia problems.
{\small
\bibliographystyle{ieee}
\bibliography{VITAL}

\begin{thebibliography}{10}\itemsep=-1pt

\bibitem{Long:Arxiv2018}
E.~S. Chengjiang~Long, Roddy~Collins and A.~Hoogs.
\newblock Deep neural networks in fully connected crf for image labeling with
  social network metadata.
\newblock In {\em IEEE Winter Conf. on Applications of Computer Vision (WACV)}.
  IEEE, 2018.

\bibitem{Crowston:2012amazon}
K.~Crowston.
\newblock Amazon mechanical turk: A research tool for organizations and
  information systems scholars.
\newblock In {\em Shaping the Future of ICT Research. Methods and Approaches}.
  2012.

\bibitem{Durugkar:Arxiv2016}
I.~Durugkar, I.~Gemp, and S.~Mahadevan.
\newblock Generative multi-adversarial networks.
\newblock {\em arXiv}, 2016.

\bibitem{guillaumin2010}
M.~Guillaumin, J.~Verbeek, and C.~Schmid.
\newblock {Multimodal semi-supervised learning for image classification}.
\newblock In {\em IEEE Conference on Computer Vision and Pattern Recognition
  (CVPR)}, 2010.

\bibitem{HeResnet:CVPR2016}
K.~He, X.~Zhang, S.~Ren, and J.~Sun.
\newblock Deep residual learning for image recognition.
\newblock In {\em IEEE Conference on Computer Vision and Pattern Recognition
  (CVPR)}, 2016.

\bibitem{Hu:CVPR2016learning}
H.~Hu, G.-T. Zhou, Z.~Deng, Z.~Liao, and G.~Mori.
\newblock Learning structured inference neural networks with label relations.
\newblock In {\em IEEE Conference on Computer Vision and Pattern Recognition
  (CVPR)}, 2016.

\bibitem{hua2013collaborative}
G.~Hua, C.~Long, M.~Yang, and Y.~Gao.
\newblock Collaborative active learning of a kernel machine ensemble for
  recognition.
\newblock In {\em IEEE International Conference on Computer Vision (ICCV)},
  2013.

\bibitem{hua2018collaborative}
G.~Hua, C.~Long, M.~Yang, and Y.~Gao.
\newblock Collaborative active visual recognition from crowds: A distributed
  ensemble approach.
\newblock {\em IEEE transactions on pattern analysis and machine intelligence},
  40(3):582--594, 2018.

\bibitem{Johnson:ICCV2015}
J.~Johnson, L.~Ballan, and L.~Fei-Fei.
\newblock Love thy neighbors: Image annotation by exploiting image metadata.
\newblock In {\em IEEE International Conference on Computer Vision (ICCV)},
  2015.

\bibitem{Johnson:NAACL15}
R.~Johnson and T.~Zhang.
\newblock Effective use of word order for text categorization with
  convolutional neural networks.
\newblock In {\em The Conference of the North American Chapter of the
  Association for Computational Linguistics (NAACL)}, 2015.

\bibitem{Kalchbrenner:ACL2014}
N.~Kalchbrenner, E.~Grefenstette, and P.~Blunsom.
\newblock A convolutional neural network for modelling sentences.
\newblock In {\em The Association for Computational Linguistics (ACL)}, 2014.

\bibitem{Kim:EMNLP2014}
Y.~Kim.
\newblock Convolutional neural networks for sentence classification.
\newblock In {\em The Conference on Empirical Methods in Natural Language
  Processing (EMNLP)}, 2014.

\bibitem{Lindstaedt2008}
S.~Lindstaedt, V.~P. adn Roland~Morzinger, R.~Kern, H.~Mulner, and C.~Wagne.
\newblock Recommending tags for pictures based on text, visual content and user
  context.
\newblock In {\em International Conference on Internet and Web Applicants and
  Services (ICIW)}, 2008.

\bibitem{Liu:CVPR2017semantic}
F.~Liu, T.~Xiang, T.~M. Hospedales, W.~Yang, and C.~Sun.
\newblock Semantic regularisation for recurrent image annotation.
\newblock In {\em IEEE Conference on Computer Vision and Pattern Recognition
  (CVPR)}, 2017.

\bibitem{Liu:CVPR2018}
N.~Liu, J.~Han, and M.-H. Yang.
\newblock Picanet: Learning pixel-wise contextual attention for saliency
  detection.
\newblock In {\em IEEE Conference on Computer Vision and Pattern Recognition
  (CVPR)}, 2018.

\bibitem{long2015multi}
C.~Long and G.~Hua.
\newblock Multi-class multi-annotator active learning with robust gaussian
  process for visual recognition.
\newblock In {\em IEEE International Conference on Computer Vision (ICCV)},
  2015.

\bibitem{Long_2017_CVPR}
C.~Long and G.~Hua.
\newblock Correlational gaussian processes for cross-domain visual recognition.
\newblock In {\em IEEE Conference on Computer Vision and Pattern Recognition
  (CVPR)}, 2017.

\bibitem{long2013active}
C.~Long, G.~Hua, and A.~Kapoor.
\newblock Active visual recognition with expertise estimation in crowdsourcing.
\newblock In {\em IEEE International Conference on Computer Vision (ICCV)},
  pages 3000--3007, 2013.

\bibitem{long2016joint}
C.~Long, G.~Hua, and A.~Kapoor.
\newblock A joint gaussian process model for active visual recognition with
  expertise estimation in crowdsourcing.
\newblock {\em International journal of computer vision (IJCV)},
  116(2):136--160, 2016.

\bibitem{McAuleyECCV12}
J.~J. McAuley and J.~Leskovec.
\newblock Image labeling on a network: Using social-network metadata for image
  classification.
\newblock In {\em The European Conference on Computer Vision (ECCV)}, 2012.

\bibitem{Nguyen:CVPR2017}
A.~Nguyen, J.~Clune, Y.~Bengio, A.~Dosovitskiy, and J.~Yosinski.
\newblock Plug \& play generative networks: Conditional iterative generation of
  images in latent space.
\newblock In {\em IEEE Conference on Computer Vision and Pattern Recognition
  (CVPR)}, 2017.

\bibitem{Nilsback:2008automated}
M.-E. Nilsback and A.~Zisserman.
\newblock Automated flower classification over a large number of classes.
\newblock In {\em Computer Vision, Graphics \& Image Processing (ICVGIP)},
  2008.

\bibitem{Niu:TIP2019multi}
Y.~Niu, Z.~Lu, J.-R. Wen, T.~Xiang, and S.-F. Chang.
\newblock Multi-modal multi-scale deep learning for large-scale image
  annotation.
\newblock {\em IEEE Transactions on Image Processing (TIP)}, 28(4):1720--1731,
  2019.

\bibitem{Plummer:ECCV2018}
B.~A. Plummer, P.~Kordas, M.~Hadi~Kiapour, S.~Zheng, R.~Piramuthu, and
  S.~Lazebnik.
\newblock Conditional image-text embedding networks.
\newblock In {\em The European Conference on Computer Vision (ECCV)}, September
  2018.

\bibitem{Reed:CVPR2016learning}
S.~Reed, Z.~Akata, H.~Lee, and B.~Schiele.
\newblock Learning deep representations of fine-grained visual descriptions.
\newblock In {\em IEEE Conference on Computer Vision and Pattern Recognition
  (CVPR)}, 2016.

\bibitem{Reed:NIPS2016}
S.~E. Reed, Z.~Akata, S.~Mohan, S.~Tenka, B.~Schiele, and H.~Lee.
\newblock Learning what and where to draw.
\newblock In {\em Advances in Neural Information Processing Systems (NIPS)},
  2016.

\bibitem{Reed:ICML2016}
S.~E. Reed, Z.~Akata, X.~Yan, L.~Logeswaran, B.~Schiele, and H.~Lee.
\newblock Generative adversarial text to image synthesis.
\newblock In {\em The International Conference on Machine Learning (ICML)},
  2016.

\bibitem{Russell:2008labelme}
B.~C. Russell, A.~Torralba, K.~P. Murphy, and W.~T. Freeman.
\newblock Labelme: a database and web-based tool for image annotation.
\newblock {\em International journal of computer vision (IJCV)}, 2008.

\bibitem{SawantDLW10}
N.~Sawant, R.~Datta, J.~Li, and J.~Z. Wang.
\newblock Quest for relevant tags using local interaction networks and visual
  content.
\newblock In {\em The {ACM} {SIGMM} International Conference on Multimedia
  Information Retrieval (MIR)}, 2010.

\bibitem{sigurbjoernsson2008}
B.~Sigurbjörnsson and R.~van Zwol.
\newblock Flickr tag recommendation based on collective knowledge.
\newblock In {\em The International Conference on World Wide Web (WWW)}, 2008.

\bibitem{Stone2008}
Z.~Stone, T.~Zickler, and T.~Darrell.
\newblock Autotagging facebook: Social network context improves photo
  annotation.
\newblock In {\em IEEE Conference on Computer Vision and Pattern Recognition
  Workshop (CVPRW)}, 2008.

\bibitem{Wah:CUB_200_2011}
C.~Wah, S.~Branson, P.~Welinder, P.~Perona, and S.~Belongie.
\newblock {The Caltech-UCSD Birds-200-2011 Dataset}.
\newblock Technical Report CNS-TR-2011-001, California Institute of Technology,
  2011.

\bibitem{Xu:CVPR2018}
T.~Xu, P.~Zhang, Q.~Huang, H.~Zhang, Z.~Gan, X.~Huang, and X.~He.
\newblock Attngan: Fine-grained text to image generation with attentional
  generative adversarial networks.
\newblock In {\em IEEE Conference on Computer Vision and Pattern Recognition
  (CVPR)}, 2018.

\bibitem{Yeh:CVPR2018}
R.~A. Yeh, M.~N. Do, and A.~G. Schwing.
\newblock Unsupervised textual grounding: Linking words to image concepts.
\newblock In {\em IEEE Conference on Computer Vision and Pattern Recognition
  (CVPR)}, 2018.

\bibitem{Yin:Arxiv2016multichannel}
W.~Yin and H.~Sch{\"u}tze.
\newblock Multichannel variable-size convolution for sentence classification.
\newblock {\em arXiv}, 2016.

\bibitem{zhang:Arxiv2017stackgan}
H.~Zhang, T.~Xu, H.~Li, S.~Zhang, X.~Huang, X.~Wang, and D.~Metaxas.
\newblock Stackgan: Text to photo-realistic image synthesis with stacked
  generative adversarial networks.
\newblock {\em arXiv preprint}, 2017.

\bibitem{zhang:Arxiv2017}
H.~Zhang, T.~Xu, H.~Li, S.~Zhang, X.~Wang, X.~Huang, and D.~Metaxas.
\newblock Stackgan++: Realistic image synthesis with stacked generative
  adversarial networks.
\newblock {\em arXiv}, 2017.

\end{thebibliography}
}

\end{document}